\documentclass[final,1p,times]{elsarticle}

\usepackage{amssymb}
\usepackage{xcolor}
\usepackage{mathtools}
\usepackage{amsmath}
\usepackage{graphicx}
\usepackage{epsfig}
\usepackage{subfigure}
\usepackage{epstopdf}
\usepackage{url}
\usepackage{multirow}
\usepackage{float}
\usepackage{caption}

\makeatletter
\renewcommand\section{\@startsection{section}{1}{\z@}%
{-2.5ex plus -1ex minus -0.5ex}{0.2ex plus 0.5ex minus 0ex}{\normalfont\normalsize\bf}}
\makeatother

\makeatletter
\renewcommand\subsection{\@startsection{subsection}{1}{\z@}%
{-1.5ex plus -1ex minus -0.5ex}{0.1ex plus 0.1ex minus 0.1ex}{\normalfont\normalsize}}

\makeatother

\makeatletter
\renewcommand\subsubsection{\@startsection{subsubsection}{1}{\z@}%
{-1ex plus -1ex minus -0.5ex}{0.1ex plus 0.1ex}{\normalfont\normalsize}}
\makeatother

\begin{document}

\begin{frontmatter}



\title{Robust Face Recognition via Block Sparse Bayesian Learning}


\author{Taiyong Li$^{1,2}$}
\author{Zhilin Zhang$^{3,4,*}$}

\address{$^1$School of Financial Information Engineering, Southwestern University of Finance and Economics, Chengdu 610074, China}
\address{$^2$Institute of Chinese Payment System, Southwestern University of Finance and Economics, Chengdu 610074, China}
\address{$^3$Department of Electrical and Computer Engineering, University of California at San Diego, La Jolla, CA 92093-0407, USA}
\address{$^4$Samsung R\&D Institute America - Dallas, 1301 East Lookout Drive, Richardson, TX 75082, USA}
\cortext[cor1]{Corresponding author: zhilinzhang@ieee.org (Zhilin Zhang) }

\begin{abstract}
Face recognition (FR) is an important task in pattern recognition and computer vision. Sparse representation (SR) has been demonstrated to be a powerful framework for FR. In general, an SR algorithm treats each face in a training dataset as a basis function, and tries to find a sparse representation of a test face under these basis functions. The sparse representation coefficients then provide a recognition hint. Early SR algorithms are based on a basic sparse model. Recently, it has been found that algorithms based on a block sparse model can achieve better recognition rates. Based on this model, in this study we use block sparse Bayesian learning (BSBL) to find a sparse representation of a test face for recognition. BSBL is a recently proposed framework, which has many advantages over existing block-sparse-model based algorithms. Experimental results on the Extended Yale B, the AR and the CMU PIE face databases show that using BSBL can achieve better recognition rates and higher robustness than state-of-the-art algorithms in most cases.
\end{abstract}

\begin{keyword}

Face Recognition \sep Classification \sep Sparse Representation \sep Sparse Learning \sep Block Sparse Bayesian Learning (BSBL) \sep Block Sparsity

\end{keyword}

\end{frontmatter}


\section{Introduction}
Owing to the rapid development of network and computer technologies, face recognition(FR) plays an important role in many applications, such as video surveillance, man-machine interface, digital entertainment and so on. Many methods of FR have been developed over the past two decades \cite{martinez2001pca,phillips2005overview,turk1991face,brunelli1993face,he2005face}. Basically, FR is a typical problem of classification.

In a typical FR system, besides the face detection and face alignment, there are two main stages in the process of FR. One is feature extraction, which obtains a set of relevant information from a face image for further classification. Because of the huge size of face images, it is desired to extract features from each face image, which have lower dimensions and facilitate recognition. Lots of feature extraction methods have been proposed, such as PCA \cite{turk1991face,turk1991eigenfaces}, LPP \cite{he2005face} and LDA \cite{belhumeur1997eigenfaces} and so on. Another stage is classification, which builds a classification model and assigns a label to a test face image. There are many classification algorithms. Typical algorithms include Nearest Neighbor (NN) \cite{duda2012pattern}, Nearest Subspace (NS) \cite{ho2003clustering} and Support Vector Machine (SVM) \cite{vapnik1999nature}.

Recently, Wright \emph{et al.} proposed a novel FR method called Sparse Representation Classification (SRC) \cite{wright2009robust}. In this method, face images in the training set form a dictionary matrix (each face image is vectorized and forms a column of the dictionary matrix), and then a vectorized test face image is represented under this dictionary matrix. The representation coefficients provide hints for recognition. For example, if a test face image and a training face image belong to the same subject, then the representation coefficients of the vectorized test face image under the dictionary matrix are sparse (or compressive), i.e., most coefficients are zero (or close to zero). For each class (i.e., the columns in the dictionary matrix which are associated with a subject), one can calculate the reconstruction error of the vectorized test face image using these columns and the associated representation coefficients. The class with the minimum reconstruction error suggests the test face image belongs to this class. More frequently, one uses a feature vector extracted from a face image, instead of the original vectorized face image, in this method. SRC is so robust that it can achieve good performance in occlusion and noise environments.

Following the idea of SRC, a number of SRC related recognition methods have been proposed. Gao \emph{et al.} extended the basic SRC method to a kernel version \cite{gao2010kernel}. Yang \emph{et al.} proposed a face recognition method via sparse coding which is much more robust than SRC in occlusion, corruption and disguise environments \cite{yang2011robust}. Some other works improved the basic SRC method using weighted sparse representations \cite{lu2012face}, Gabor feature based sparse representations \cite{yang2010gabor}, dimensionality reduction \cite{zhang2010dimensionality}, locally adaptive sparse representations \cite{chen2010robust} and supervised sparse representations \cite{xu2011supervised}.

Recently, it is found that using algorithms based on a block sparse model \cite{grouplasso}, instead of the algorithms based on the basic sparse representation model, can achieve higher recognition rates in face recognition \cite{elhamifar2012block}. However, these algorithms ignore intra-block correlation in representation coefficients. The existence of intra-block correlation in representation coefficients results from the fact that training face images with the same class as a test face image are all correlated with the test face image, and thus the representation coefficients associated with the training face images are not independent. In sparse reconstruction scenarios it is shown that exploiting the intra-block correlation can significantly improve algorithmic performance \cite{Zhang2012TSP}.

In this study, we use block sparse Bayesian learning (BSBL) \cite{Zhang2012TSP} to estimate the  representation coefficients. BSBL has many advantages over existing block-sparse-model based algorithms, especially it has the ability to exploit the intra-block correlation in representation coefficients for better algorithmic performance. Experimental results on the Extended Yale B ,the AR and the CMU PIE databases show that BSBL achieves better results than state-of-the-art SRC algorithms in most cases.

The rest of this paper is organized as follows. Section \ref{sec:relatedWork} gives a brief review of the original face recognition via sparse representation. Section \ref{sec:SBLandBSBL} introduces sparse Bayesian learning. The Block Sparse Bayesian Learning approach for face recognition is proposed in Section \ref{sec:FRBSBL}. Experimental results are reported in Section \ref{sec:experiments}. Conclusion is drawn in the last section.

\section{Related work}
\label{sec:relatedWork}

\subsection{Face recognition via sparse representation}
We first describe the basic SRC method \cite{wright2009robust} for face recognition. Given training faces of all \emph{K} subjects, a dictionary matrix is formed as follows
\begin{equation}
\mathbf{\Phi}\triangleq[\mathbf{\Phi}_1,\mathbf{\Phi}_2,...,\mathbf{\Phi}_K]
\label{eq1}
\end{equation}
where $\mathbf{\Phi}_i=[\mathbf{v}_{i,1},\mathbf{v}_{i,2},\ldots,\mathbf{v}_{i,n_i}]\in\mathbb{R}^{m\times{n_i}}$, and $\mathbf{v}_{i,j}$ is the $j$-th face \footnote{For simplicity, we describe $\mathbf{v}_{i,j}$ as a vectorized face image. But in practice, $\mathbf{v}_{i,j}$ is a feature vector extracted from the face image, as done in our experiments.} of the $i$-th subject. Then, a vectorized test face $\mathbf{y} \in \mathbb{R}^{m\times 1}$ is represented under the dictionary matrix as follows
\begin{eqnarray}
\mathbf{y} &=& \mathbf{\Phi} \mathbf{x}  \nonumber \\
&=& \mathbf{v}_{1,1}  x_{1,1} + \cdots +  \mathbf{v}_{1,n_1} x_{1,n_1} + \cdots +   \mathbf{v}_{i-1,n_{i-1}} x_{i-1,n_{i-1}}      \nonumber \\
&& + \mathbf{v}_{i,1} x_{i,1}  +  \mathbf{v}_{i,2} x_{i,2} + \cdots + \mathbf{v}_{i,n_i} x_{i,n_i} \nonumber \\
&& +  \mathbf{v}_{i+1,1} x_{i+1,1} + \cdots +  \mathbf{v}_{K,n_K} x_{K,n_K}
\label{equ:sparseRepresentation}
\end{eqnarray}
where $\mathbf{x} \triangleq [x_{1,1}, \cdots, x_{i,1},\cdots,x_{i,n_i},\cdots,x_{K,n_K}]^T$ is the representation coefficient vector. In the basic SRC method, it is suggested that if the new test face $\mathbf{y}$ belongs to a subject in the training set, say the $i$-th subject, then under a sparsity constraint on $\mathbf{x}$, only some of the coefficients $x_{i,1}, x_{i,2}, \cdots, x_{i,n_i}$ are significantly nonzero, while other coefficients, i.e. $x_{j,k}\, (j\neq i,\forall k)$, are zero or close to zero.

Mathematically, the above idea can be described as the following sparse representation problem
\begin{equation}
\widehat{\mathbf{x}}_0 = \arg \min_\mathbf{x} \|\mathbf{x}\|_0 \quad \mathrm{s.t}. \quad \mathbf{y} =\mathbf{\Phi} \mathbf{x}
\label{eq4}
\end{equation}
where $\|\mathbf{x}\|_0$ counts the number of nonzero elements in the vector $\mathbf{x}$. Once we have obtained the solution $\widehat{\mathbf{x}}_0$, the class label of $\mathbf{y}$ can be found by
\begin{equation}
i = \arg \min_j \|\mathbf{y} - \mathbf{\Phi} \delta_j(\widehat{\mathbf{x}}_0)\|_2,
\label{eq6}
\end{equation}
where $\delta_j(\mathbf{x}):\mathbb{R}^n\rightarrow\mathbb{R}^n$ is the characteristic function which maintains the elements of $\mathbf{x}$ associated with the $j$-th class, while sets other elements of $\mathbf{x}$ to zero.

However, finding the solution to (\ref{eq4}) is NP-hard \cite{natarajan1995sparse}. Recent theories in compressed sensing \cite{donoho2006most,candes2006near} show that if the true solution is sparse enough, under some mild conditions the solution can be found by solving the following convex-relaxation problem
\begin{equation}
\widehat{\mathbf{x}}_1 = \arg \min_\mathbf{x} \|\mathbf{x}\|_1 \quad \mathrm{s.t}. \quad \mathbf{y}=\mathbf{\Phi}\mathbf{x}.
\label{eq5}
\end{equation}
Further, to deal with small dense model noise, the problem (\ref{eq5}) can be changed to the following one
\begin{equation}
\widehat{\mathbf{x}}_1 = \arg \min_\mathbf{x} \|\mathbf{x}\|_1 \quad \mathrm{s.t}. \quad \| \mathbf{y} - \mathbf{\Phi}\mathbf{x} \|_2 \leq \epsilon
\label{eq7}
\end{equation}
where $\epsilon$ is a noise-tolerance constant. Many $\ell_1$-minimization algorithms can be used to find the solution to (\ref{eq5}) or to (\ref{eq7}), such as LASSO \cite{lasso} and Basis Pursuit Denoising \cite{BP}.

In a practical face recognition problem, the coefficient vector $\widehat{\mathbf{x}}_1$ (or $\widehat{\mathbf{x}}_0$) is not only sparse but also block sparse. To see this, we can rewrite the sparse representation problem (\ref{equ:sparseRepresentation}) as follows
\begin{eqnarray}
\mathbf{y} = \mathbf{\Phi} \mathbf{x} = [\mathbf{\Phi}_1,\mathbf{\Phi}_2,...,\mathbf{\Phi}_K] \mathbf{x} = \sum_{j=1}^K \mathbf{\Phi}_j \mathbf{x}_j
\label{equ:blockSparseRepresentation}
\end{eqnarray}
where $\mathbf{x}_j \in \mathbb{R}^{n_j \times 1}$ is the coefficient vector associated with the $j$-th class, and $\mathbf{x} \triangleq [\mathbf{x}_1^T,\cdots,\mathbf{x}_K^T]^T$. When a test face $\mathbf{y} $ belongs to the $j$-th class, ideally only elements in $\mathbf{x}_j$ are significantly nonzero. In other words, only the block $\mathbf{x}_j$ has significantly nonzero norm. Clearly, this is a canonical block sparse model \cite{grouplasso,ModelCS}. Many algorithms for the block sparse model can be used here. For example, in \cite{elhamifar2012block} it is suggested  to use the following algorithm:
\begin{equation}
\widehat{\mathbf{x}}_{2,1} = \arg \min_\mathbf{x} \sum_{j=1}^K \|\mathbf{x}_j\|_2 \quad \mathrm{s.t}. \quad \| \mathbf{y} - \mathbf{\Phi}\mathbf{x} \|_2 \leq \epsilon
\label{equ:blockL1}
\end{equation}
This is a natural extension of basic $\ell_1$-minimization algorithms, which imposes $\ell_2$ norm on block elements and then $\ell_1$ norm over blocks. It has been shown that exploiting the block structure can largely improve the estimation quality of $\widehat{\mathbf{x}}_0$ \cite{ModelCS,huang2010benefit,rao2012universal}.

However, one should note that when the test face belongs to the $j$-th class, not only the representation coefficient block $\mathbf{x}_j$ is a nonzero block, but also its elements are correlated in amplitude. The correlation arises because the faces of the $j$-th class in the training set are all correlated with the test face, and thus the elements in $\mathbf{x}_j$ are mutually dependent. It is shown that exploiting the correlation within blocks can further improve the estimation quality of $\widehat{\mathbf{x}}_0$ \cite{Zhang2012TSP,Zhang2012TBME} than only exploiting the block structure.

Therefore, in this study we propose to use block sparse Bayesian learning (BSBL) \cite{Zhang2012TSP} to estimate $\widehat{\mathbf{x}}_0$ by exploiting the block structure and the correlation within blocks. In the next section we first briefly introduce sparse Bayesian learning (SBL), and then introduce BSBL.

\section{SBL and BSBL}
\label{sec:SBLandBSBL}

SBL \cite{tipping2001sparse} was initially proposed as a machine learning method. But later it has been shown to be a powerful method for sparse representation, sparse signal recovery and compressed sensing.

\subsection{Advantages of SBL}
\label{subsec:SBL}

Compared to LASSO-type algorithms (such as the original LASSO algorithm, Basis Pursuit Denoising, Group Lasso, Group Basis Pursuit, and other algorithms based on $\ell_1$-minimization), SBL has the following advantages \cite{ZhangDissertation,Wipf2006Thesis}.

\begin{enumerate}

  \item Its recovery performance is robust to the characteristics of the matrix $\mathbf{\Phi}$, while most other algorithms are not. For example, it has been shown that when columns of  $\mathbf{\Phi}$ are highly coherent, SBL still maintains good performance, while algorithms such as LASSO have seriously degraded performance \cite{Wipf2011NIPS}. This advantage is very attractive to sparse representation and other applications, since in these applications the matrix $\mathbf{\Phi}$ is not a random matrix and its columns are highly coherent.

   \item SBL has a number of desired advantages over many popular algorithms in terms of local and global convergence. It can be shown that SBL provides a sparser solution than Lasso-type algorithms. In particular, in noiseless situations and under certain conditions, the global minimum of SBL cost function is unique and corresponds to the true sparsest solution, while the global minimum of the cost function of LASSO-type algorithms is not necessarily the true sparsest solution \cite{Wipf2004IEEE}. These advantages imply that SBL is a better choice in feature selection via sparse representation \cite{Jing2012CVPR}.

    \item Recent works in SBL \cite{Zhang2012TSP,Zhang2011IEEE} provide robust learning rules for automatically estimating values of its regularizer (related to noise variance) such that SBL algorithms can achieve good performance. In contrast, LASSO-type algorithms generally need users to choose values for such regularizer, which is often obtained by cross-validation. However, this takes lots of time for large-scale datasets, which is not convenient and even impossible in some scenarios.

\end{enumerate}

\subsection{Introduction to BSBL}

BSBL \cite{Zhang2012TSP} is an extension of the basic SBL framework, which exploits a block structure and intra-block correlation in the coefficient vector $\mathbf{x}$. It is based on the assumption that $\mathbf{x}$ can be partitioned into $K$ non-overlapping blocks:
\begin{eqnarray}
\mathbf{x} = [ \underbrace{x_{1,1},\cdots,x_{1,n_1}}_{\mathbf{x}_1^T},   \cdots,  \underbrace{x_{K,1},\cdots,x_{K, n_K}}_{\mathbf{x}_K^T}]^T
\label{equ:partition}
\end{eqnarray}
Among these blocks, few blocks are nonzero. Then, each block $\mathbf{x}_i \in \mathbb{R}^{n_i \times 1}$ is assumed to satisfy a parameterized multivariate Gaussian distribution:
\begin{eqnarray}
p(\mathbf{x}_i; \gamma_i, \mathbf{B}_i) \sim  \mathcal{N}(\textbf{0},\gamma_i \mathbf{B}_i), \quad i=1,\cdots,K
\end{eqnarray}
with the unknown parameters $\gamma_i$ and $\mathbf{B}_i$. Here $\gamma_i$ is a nonnegative parameter controlling the block-sparsity of $\mathbf{x}$. When $\gamma_i=0$, the $i$-th block becomes zero. During the learning procedure most $\gamma_i $ tend to be zero, due to the mechanism of automatic relevance determination \cite{tipping2001sparse}. Thus sparsity at the block level is encouraged. $\mathbf{B}_i \in \mathbb{R}^{d_i \times d_i}$ is a positive definite and symmetrical matrix, capturing the intra-block correlation of the $i$-th block. Under the assumption that blocks are mutually uncorrelated, the prior of $\mathbf{x}$ is $p(\mathbf{x};\{\gamma_i,\mathbf{B}_i\}_i) \sim  \mathcal{N}(\textbf{0},\mathbf{\Sigma}_0)$, where $\mathbf{\Sigma}_0 = \mathrm{diag}\{\gamma_1 \mathbf{B}_1,\cdots,\gamma_K \mathbf{B}_K\}$ \footnote{Here $\Sigma_0 = diag\{\gamma_1 B1,..., \gamma_k Bk\}$ means $\Sigma_0$ is a block diagonal matrix, and its principal diagonal blocks are $\gamma_1 B_1, ..., \gamma_k B_k$.}. To avoid overfitting, all the $\mathbf{B}_i$ will be  imposed by some constraints and their estimates will be further regularized. The model noise $\mathbf{n} \triangleq \mathbf{y}-\mathbf{\Phi} \mathbf{x}$ is assumed to satisfy  $p(\mathbf{n};\lambda) \sim  \mathcal{N}(\textbf{0},\lambda \mathbf{I})$, where $\lambda$ is a positive scalar to be estimated. Based on the above probability models, one can obtain a close-form posterior. Therefore, the estimate of $\mathbf{x}$ can be obtained by using the Maximum-A-Posteriori (MAP) estimation, providing all the parameters $\lambda, \{\gamma_i, \mathbf{B}_i\}_{i=1}^K$ are estimated.

To estimate the parameters $\lambda, \{\gamma_i, \mathbf{B}_i\}_{i=1}^K$, one can use the Type II maximum likelihood method \cite{Mackay1992evidence,tipping2001sparse}. This is equivalent to minimizing the following cost function
\begin{eqnarray}
\mathcal{L}(\Theta) &\triangleq & -2 \log \int p(\mathbf{y}|\mathbf{x};\lambda) p(\mathbf{x};\{\gamma_i,\mathbf{B}_i\}_i) d \mathbf{x}  \nonumber \\
 &=& \log|\lambda \mathbf{I} + \mathbf{\Phi} \mathbf{\Sigma}_0 \mathbf{\Phi}^T  |  + \mathbf{y}^T (\lambda \mathbf{I} + \mathbf{\Phi} \mathbf{\Sigma}_0 \mathbf{\Phi}^T)^{-1} \mathbf{y},
\label{equ:costfunc}
\end{eqnarray}
where $\Theta$ denotes all the parameters, i.e., $\Theta \triangleq \{\lambda,\{\gamma_i,\mathbf{B}_i\}_{i=1}^K \}$. There are several optimization methods to minimize the cost function, such as the expectation-maximum method, the bound-optimization method, the duality method and so on. This framework is called the BSBL framework.

BSBL not only has the advantages of the basic SBL listed in Section \ref{subsec:SBL}, but also has another two advantages:
\begin{enumerate}

  \item BSBL provides large flexibility to model and exploit correlation structure in signals, such as intra-block correlation \cite{Zhang2012TSP,Zhang2012TBME}. By exploiting the correlation structures, recovery performance is significantly improved.

  \item BSBL has the unique ability to find less-sparse \cite{Zhang2012Letter} and non-sparse \cite{Zhang2012TBME} true solutions with very small errors \footnote{Note that for an underdetermined inverse problem, i.e, $\mathbf{y}=\mathbf{\Phi} \mathbf{x}$, where $\mathbf{\Phi} \in \mathbb{R}^{m \times P}$ is one matrix or a product of a sensing matrix and a dictionary matrix as used in compressed sensing, one cannot find the true solution without any error, if the true solution $\mathbf{x}$ is non-sparse (i.e., $\|\mathbf{x}\|_0 > m$).}. This is attractive for practical use, since in practice the true solutions may not be very sparse, and existing sparse signal recovery algorithms generally fail in this case.

\end{enumerate}

Therefore, BSBL is promising for pattern recognition. In the following we use BSBL for face recognition. Among a number of BSBL algorithms, we choose the bound-optimization based BSBL algorithm \cite{Zhang2012TSP}, denoted by BSBL-BO \footnote{The BSBL-BO code can be downloaded at \url{http://dsp.ucsd.edu/~zhilin/BSBL.html}).}.

\section{Face recognition via BSBL}
\label{sec:FRBSBL}
As stated in Section 2, we use BSBL-BO to estimate $\widehat{\mathbf{x}}_0$, denoted by $\widehat{\mathbf{x}}_{\mathrm{BSBL}}$, and then use the rule (\ref{eq6}) to assign a test face $\mathbf{y}$ to a class.

In practice, a test face $\mathbf{y}$ may contain some outliers, i.e., $\mathbf{y}=\mathbf{y}_0 + \boldsymbol{\epsilon}$, where $\mathbf{y}_0$ is the outlier-free face image and $\boldsymbol{\epsilon}$ is a vector whose each entry is an outlier. Generally, the number of outliers is small, and thus $\boldsymbol{\epsilon}$ is sparse. Addressing the outlier issue is important to a practical face recognition system. In \cite{wright2009robust}, an augmented sparse model was used to deal with this issue. We now extend this method to our block sparse model, and use BSBL-BO to estimate the solution. In particular, we adopt the following augmented block sparse model:
\begin{eqnarray}
\mathbf{y} &=& \mathbf{y}_0 + \boldsymbol{\epsilon} = \mathbf{\Phi} \mathbf{x}  + \mathbf{n} + \boldsymbol{\epsilon}  \nonumber \\
&=& [\mathbf{\Phi}, \mathbf{I}] [ \mathbf{x}^T, \boldsymbol{\epsilon}^T]^T + \mathbf{n} \nonumber \\
&=& \overline{\mathbf{\Phi}} \overline{\mathbf{x}} + \mathbf{n}
\label{equ:blockL1_outlier}
\end{eqnarray}
where $\mathbf{n}$ is a vector modeling dense Gaussian noise, $\overline{\mathbf{\Phi}} \triangleq [\mathbf{\Phi}, \mathbf{I}]$ and ${\overline{\mathbf{x}}} \triangleq [\mathbf{x}^T, \boldsymbol{\epsilon}^T]^T$. Here $\mathbf{I}$ is an identity matrix of the dimension $m \times m$. Clearly, ${\overline{\mathbf{x}}}$ is also a block sparse vector, whose first $K$ blocks are the blocks of $\mathbf{x}$ and last $m$ elements are $m$ blocks with the block size being 1 \footnote{In experiments we found that treating the $m$ elements as one big block resulted in similar performance, while significantly sped up the algorithm.}. Thus, (\ref{equ:blockL1_outlier}) is still a block sparse model, and can be solved by BSBL-BO. Once BSBL-BO obtains the solution, denoted by $\widehat{\mathbf{x}}_{\mathrm{BSBL}}^{\boldsymbol{\epsilon}}$, its first $K$ blocks (denoted by $\widehat{\mathbf{x}}_{\mathrm{BSBL}}$) and its last $m$ elements (denoted by $\widehat{\boldsymbol{\epsilon}}$) are used to assign $\mathbf{y}$ to a class according to
\begin{equation}
i = \arg \min_j \|\mathbf{y} - \widehat{\boldsymbol{\epsilon}} - \mathbf{\Phi} \delta_j(\widehat{\mathbf{x}}_{\mathrm{BSBL}})\|_2
\end{equation}

We now take the Extended Yale B database \cite{GeBeKr01} as an example to show how our method works. As shown in SRC \cite{wright2009robust}, we randomly select half of the total 2414 faces (i.e., 1207 faces) as the training set and the rest as the testing set. Each face is downsampled from 192 $\times$ 168 to 24 $\times$ 21 = 504. The training set contains 38 subjects. Each subject has about 32 faces. Therefore, in our model $K=38$, and $n_1\approx\cdots\approx{n_K}\approx{32}$. The matrix $\mathbf{\Phi}$ has the size $ 504 \times 1207$, and thus the matrix $\overline{\mathbf{\Phi}}$ has the size $504 \times 1711$.

\begin{figure}[tbp]
\centering
\includegraphics[width=\textwidth]{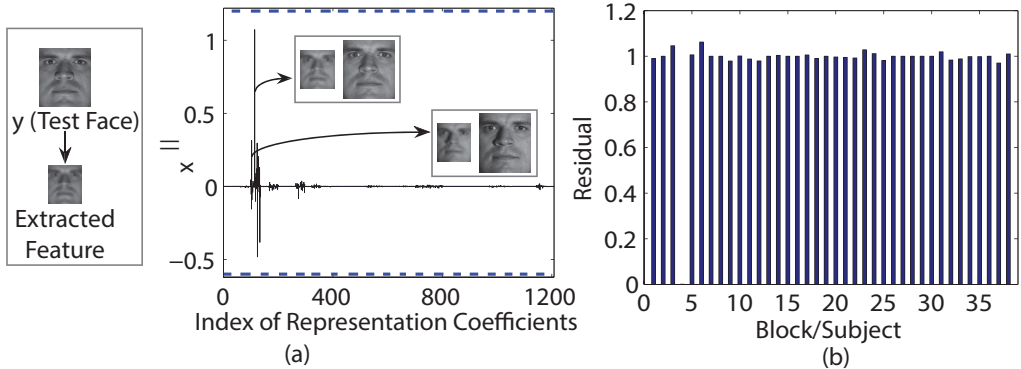}
\caption{Face recognition via BSBL. (a) Recognition with 24 $\times$ 21 downsampled faces as features. The left picture shows a test face image and the downsampled face. The right picture shows the estimated coefficients $\widehat{\mathbf{x}}_{\mathrm{BSBL}}$. The test face belongs to Subject 4, and thus the representation coefficients associated with the downsampled training faces of the subject, i.e., $\mathbf{\Phi}_4$, have large values. Two training faces (and their downsampled faces) associated with the two largest coefficients are plotted. The bars near the top and the bottom of the box indicate the blocks in the coefficient vector. (b) The residuals $\|\mathbf{y} - \mathbf{\Phi} \delta_j(\widehat{\mathbf{x}}_{\mathrm{BSBL}})\|_2$ for $j=1,\cdots, 38$.}
\label{fig:example1}
\end{figure}

The procedure is illustrated in Fig. \ref{fig:example1}. Fig.~\ref{fig:example1} (a) shows that a test face (belonging to Subject 4) can be linearly combined by a few training faces. Most of the coefficients estimated by BSBL-BO (i.e., $\widehat{\mathbf{x}}_{\mathrm{BSBL}}$) are zero or near zero and only those associated with the test face are significantly nonzero. Fig.~\ref{fig:example1} (b) shows the residuals $\|\mathbf{y} - \mathbf{\Phi} \delta_j(\widehat{\mathbf{x}}_{\mathrm{BSBL}})\|_2$ for $j=1,\cdots, 38$. The residual at $j=4$ is 0.0008, while the residuals at $j \neq 4$ are all close to 1, which makes it easy to assign the test face  to Subject 4. See Section \ref{YaleBExp} for more details.

\section{Experimental results}
\label{sec:experiments}

To demonstrate the superior performance of BSBL, we performed experiments on three widely used face databases: Extended Yale B \cite{GeBeKr01}, AR \cite{martinez1998ar} and CMU PIE \cite{sim2002cmu} face databases. The face images of these three databases were captured under varying lighting, pose or facial expression. The AR database also has occluded face images for the test of robustness of face recognition algorithms. Section \ref{subsec:withoutocclusion} shows experimental results on face images without occlusion, and Section \ref{subsec:withocclusion} shows experimental results on face images with three kinds of occlusion.

\subsection{Face recognition without occlusion}
\label{subsec:withoutocclusion}

For the experiments on face images without occlusion, we used downsampling, Eigenfaces \cite{turk1991face,turk1991eigenfaces}, and Laplacicanfaces \cite{he2005face} to reduce the dimensionality of original faces. We compared our method with three classical methods, including Nearest Neighbor (NN) \cite{duda2012pattern}, Nearest Subspace (NS) \cite{ho2003clustering}, and Support Vector Machine(SVM) \cite{vapnik1999nature}. We also compared our method with recently proposed sparse-representation based classification methods, including the basic sparse-representation classifier (SRC) \cite{wright2009robust} and the block-sparse recovery algorithm via convex optimization (BSCO) \cite{elhamifar2012block}. For NS, the subspace dimension was fixed to 9. For BSCO, we used the $P^{\prime}_{\ell_2 / \ell_1}$ algorithm \cite{elhamifar2012block} which has been shown to be the best one among all the structured sparsity-based classifiers proposed in that work.

\subsubsection{Extended Yale B database}
\label{YaleBExp}

\begin{figure}[t]
\centering
\includegraphics[width=\textwidth]{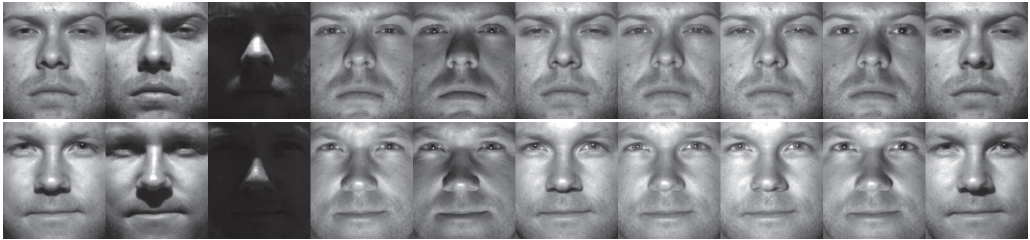}
\caption{Sample face images of 2 individuals from the Extended Yale B database. 1st row: ten sample face images of the first subject. 2nd row: ten sample face images of the third subject.}
\label{fig:ExtendedYaleB-samples}
\end{figure}

The Extended Yale B database \cite{GeBeKr01} consists of 2414 frontal-face images of 38 subjects (each subject has about 64 images). In the experiment, we used the cropped  $192 \times 168$  face images which were captured under various lighting conditions \cite{KCLee05}. Two  subjects are shown in Fig.~\ref{fig:ExtendedYaleB-samples} for illustration (for each subject, only 10 face images are shown). We randomly selected half face images of each subject as the training set and the rest as the testing set. We used downsampling, Eigenfaces, and Laplacicanfaces to extract features from face images. The dimensions of extracted features were 30, 56, 120 and 504 respectively.

\begin{figure}[!th]
	\begin{center}
	\subfigure[]{
	\epsfig{figure=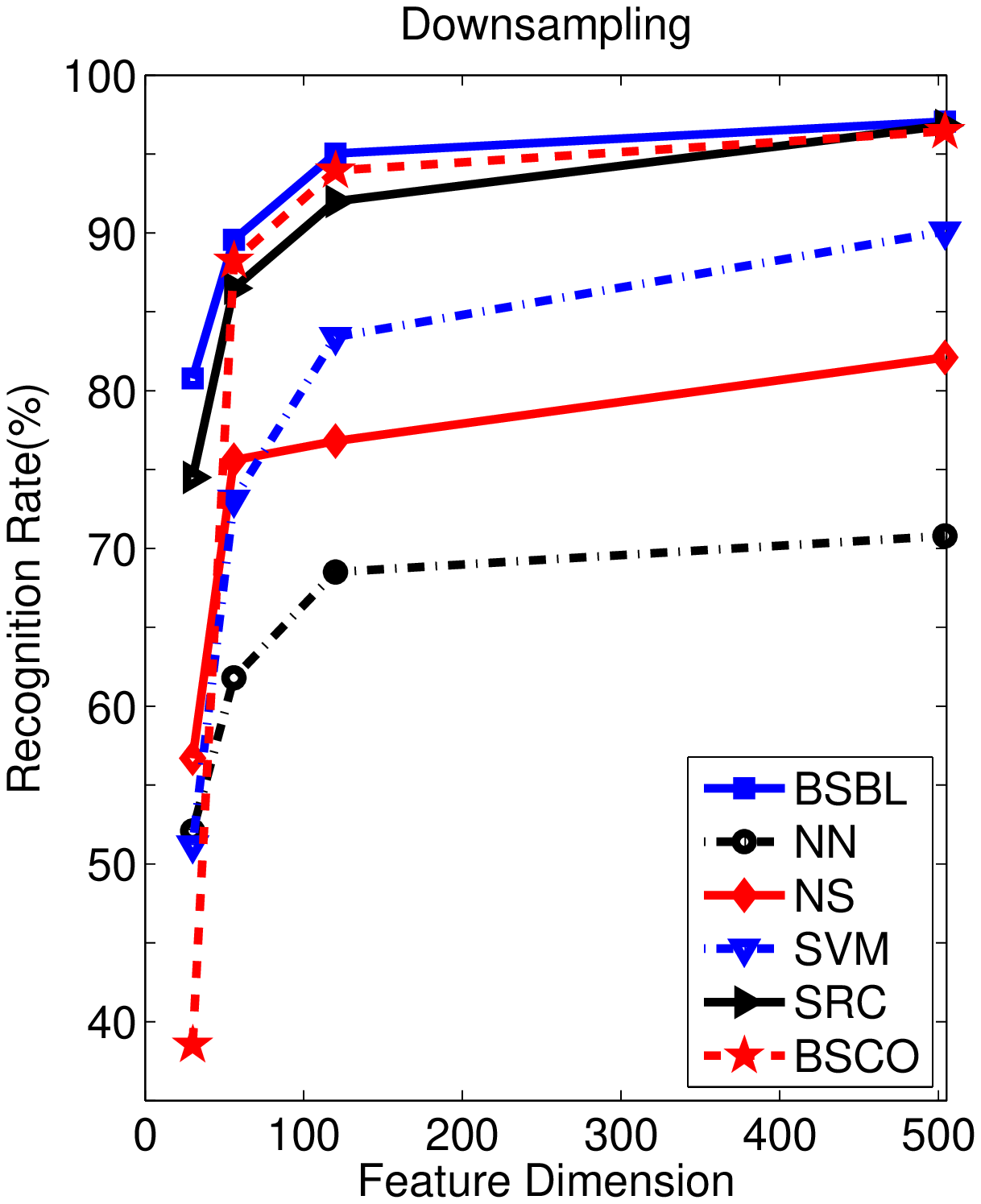,width=6cm,height=7.2cm}}
	\subfigure[]{
	\epsfig{figure=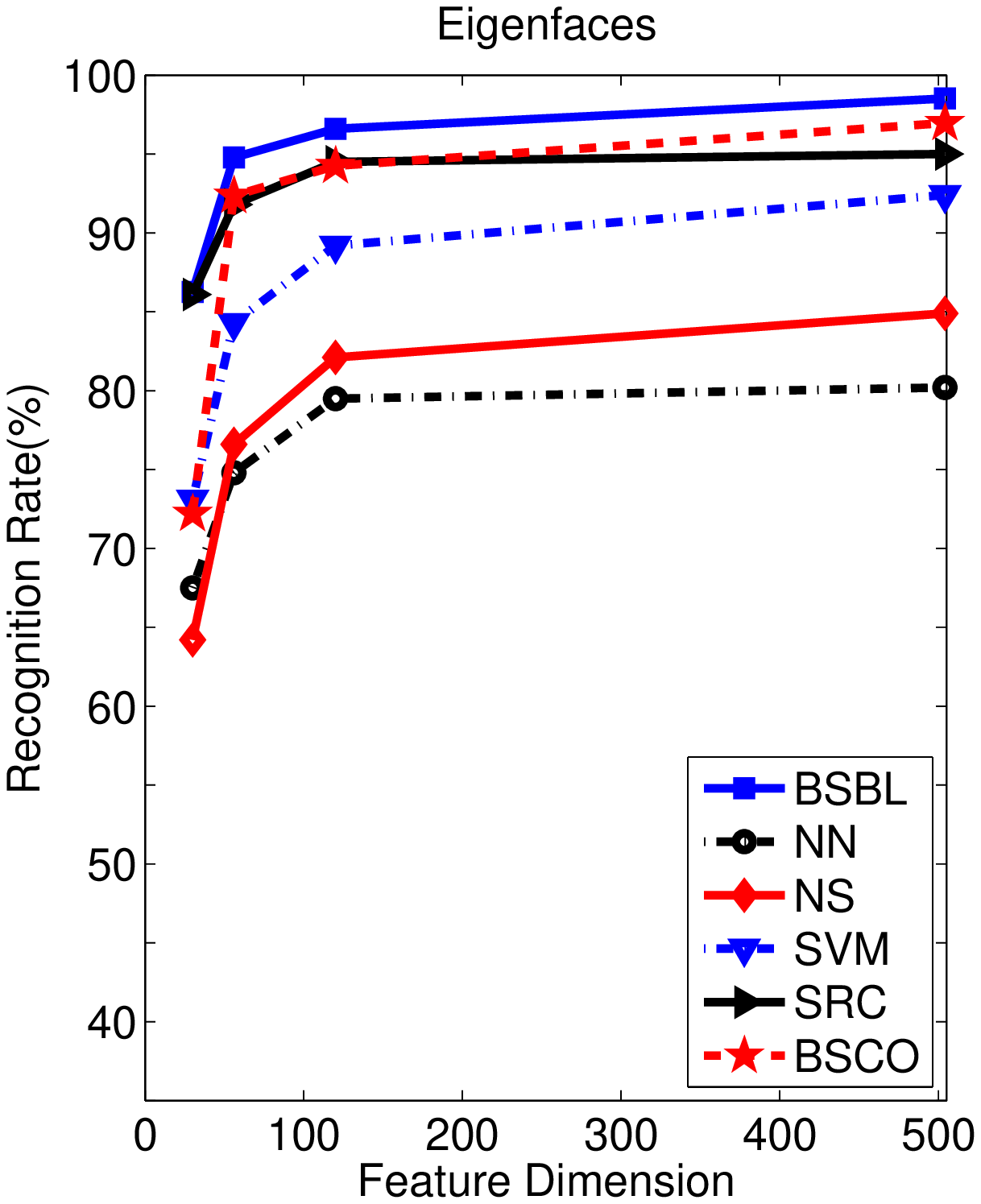,width=6cm,height=7.2cm}}
	\subfigure[]{
	\epsfig{figure=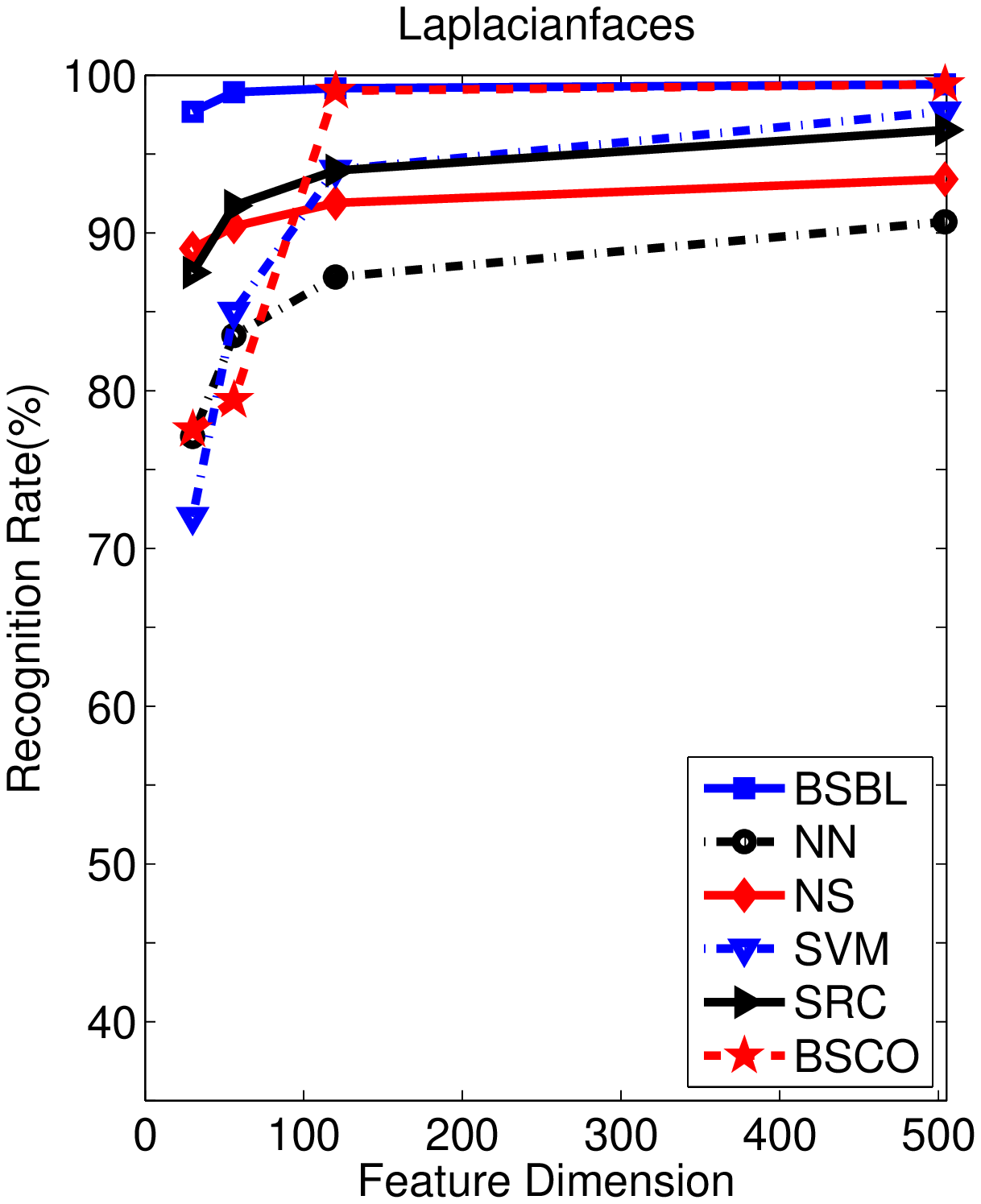,width=6cm,height=7.2cm}}
	\end{center}
	\caption{Comparison of recognition rates on Extended Yale B database when using different face features. (a) Downsampling faces. (b) Eigenfaces. (c) Laplacianfaces.}
	\label{fig:exp-yaleB}
\end{figure}

Experimental results are shown in Fig.~\ref{fig:exp-yaleB}, where we can see our method uniformly outperformed other algorithms regardless of used features. Particularly, our method had better performance when using Laplacianfaces. The superiority of our method was much clearer when the feature dimension was smaller and Laplacianfaces were used. For example, when the feature dimension was 56, our method achieved the highest rate of 98.9\%, while NN, NS, SVM, SRC and BSCO  achieved the rate of 83.5\%, 90.4\%, 85.0\%, 91.7\% and 79.4\%, respectively. Higher performance using low dimensional features is attractive for recognition, since lower feature dimension generally implies the computational load is accordingly lower.

\subsubsection{AR database}

\begin{figure}[t]
\setlength{\belowcaptionskip}{10pt}
\centering
\includegraphics[width=\textwidth]{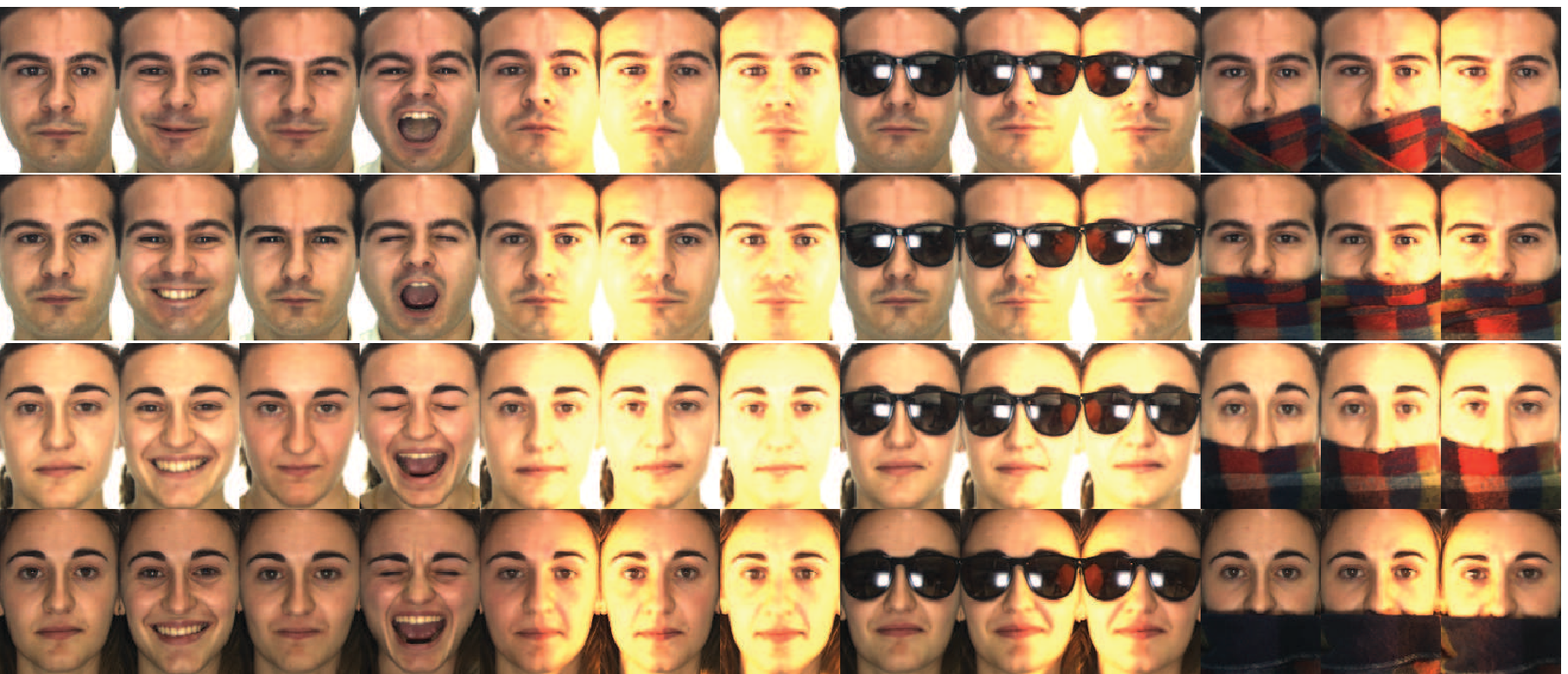}
\caption{Face images of two individuals from AR database. The 1st row: face images of the first male subject in Session 1. The 2nd row: face images of the first male subject in Session 2. The 3rd row: face images of the first female subject in Session 1. The 4th row: face images of the first female subject in Session 2.}
\label{fig:AR-samples}
\end{figure}

\begin{figure}[!th]
	\begin{center}
	\subfigure[]{
	\epsfig{figure=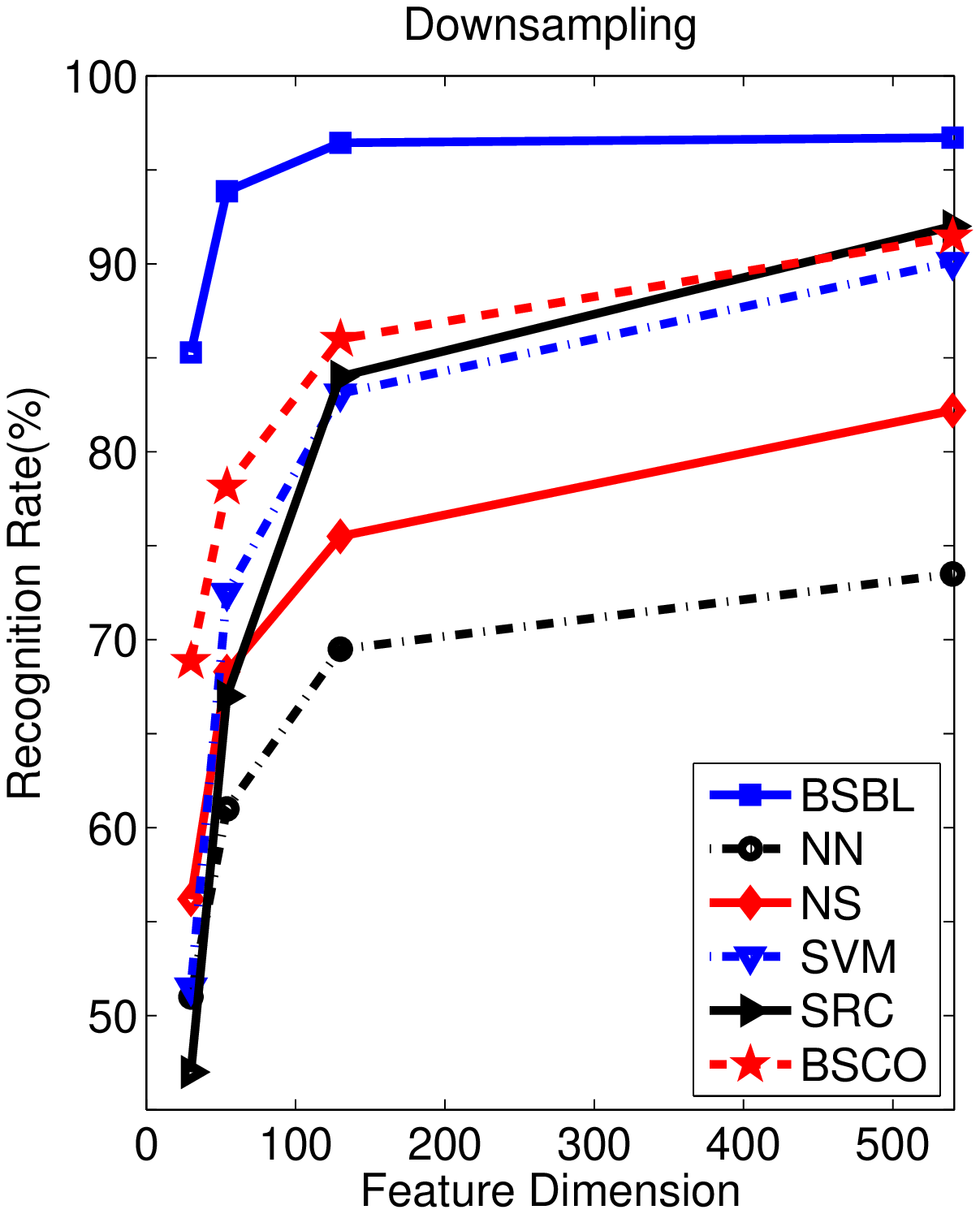,width=6cm,height=7.2cm}}
	\subfigure[]{
	\epsfig{figure=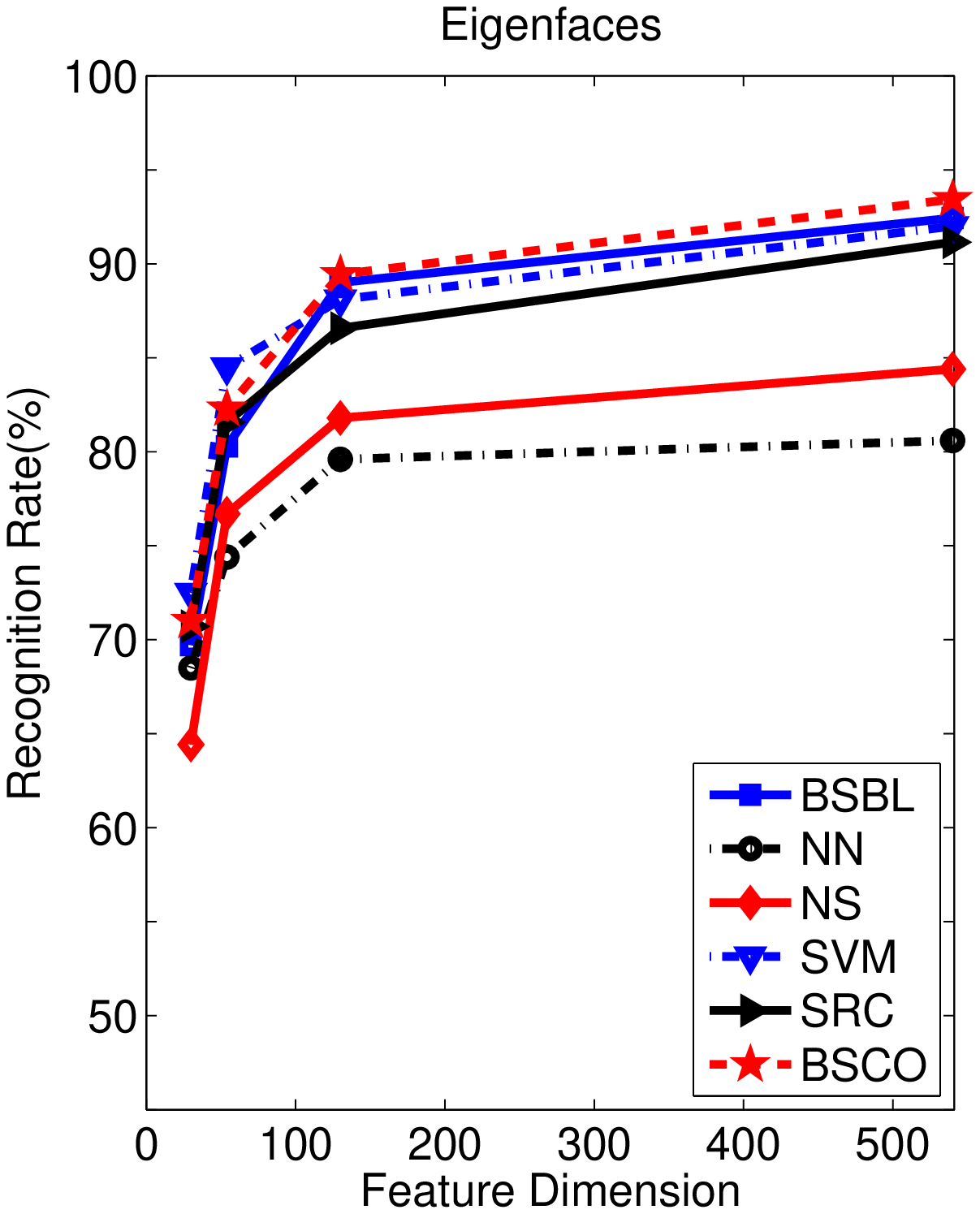,width=6cm,height=7.2cm}}
	\subfigure[]{
	\epsfig{figure=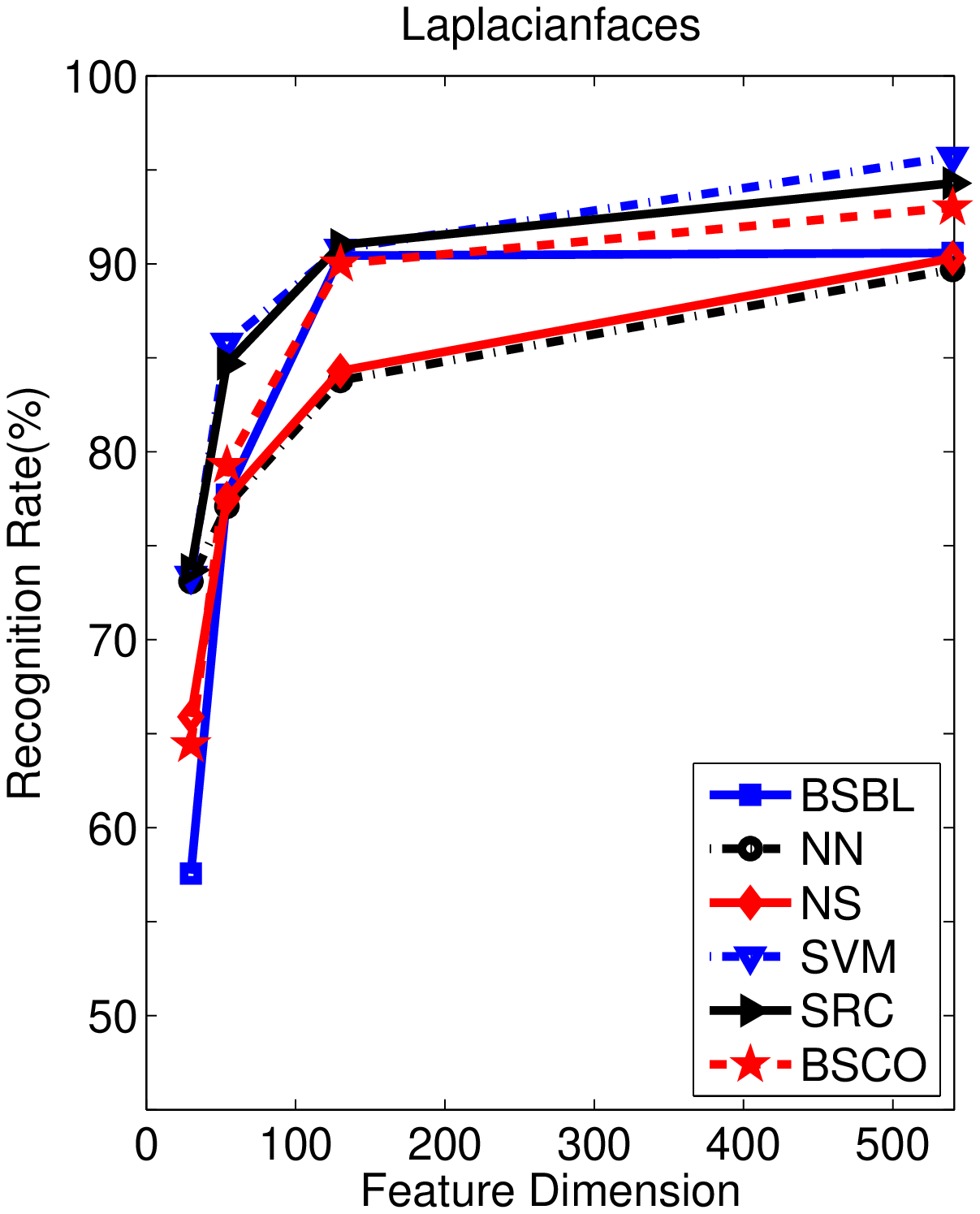,width=6cm,height=7.2cm}}
	\end{center}
	\caption{Comparison of recognition rates on AR database when using different face features. (a) Downsampling faces. (b) Eigenfaces. (c) Laplacianfaces.}
	\label{fig:exp-AR}
\end{figure}

The AR database \cite{martinez1998ar} consists of more than 4000 front-face images of 126 human subjects. Each subject has 26 images in two separated sessions, as shown in Fig.~\ref{fig:AR-samples}. This database includes more facial expression and facial disguise. We chose 100 subjects (50 male and 50 female) in this experiment. For each subject, seven face images with different illumination and facial expression (i.e., the first 7 images of each subject) in Session 1 were selected for training, and the first 7 images of each subject in Session 2 for testing. All the images were converted to gray mode and were resized to $165 \times 120$. Downsampled faces, Eigenfaces and Laplacianfaces were applied with the dimension of 30, 54, 130 and 540. Experimental results are shown in Fig.~\ref{fig:exp-AR}.

From Fig.~\ref{fig:exp-AR}(a), we can see that our algorithm significantly outperformed other classifiers when using downsampled features. However, our method did not achieve the highest rate when using Eigenfaces and Laplacianfaces. This might be due to the small block size in this experiment ($n_1=n_2=\cdots=n_{100}=7$). Although our method did not uniformly outperform other algorithms when using different face features, the recognition rate achieved by our method using downsampled faces (96.7\%) was not exceeded by other algorithms using any face features.

\subsubsection{CMU PIE database}
\label{CMUPIEExp}

\begin{figure}[t]
\setlength{\belowcaptionskip}{10pt}
\centering
\includegraphics[width=\textwidth]{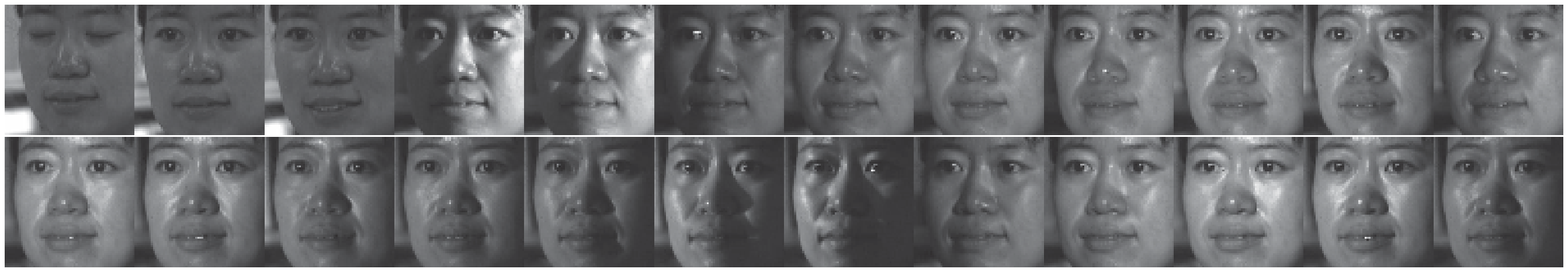}
\caption{Face images of the first individual from CMU PIE (C29) database.}
\label{fig:CMUPIE-samples}
\end{figure}

\begin{figure}[!th]
	\begin{center}
	\subfigure[]{
	\epsfig{figure=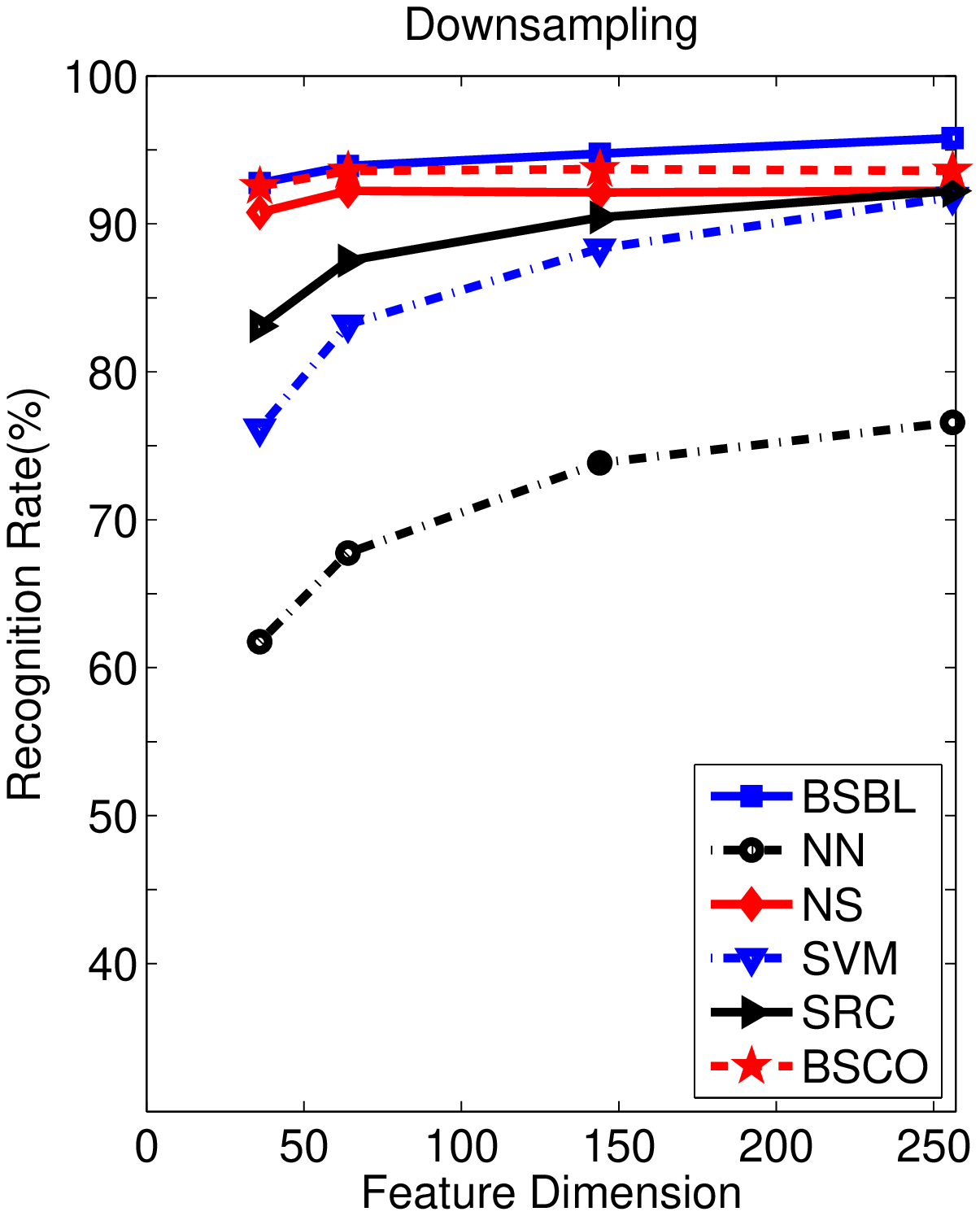,width=6cm,height=7.2cm}}
	\subfigure[]{
	\epsfig{figure=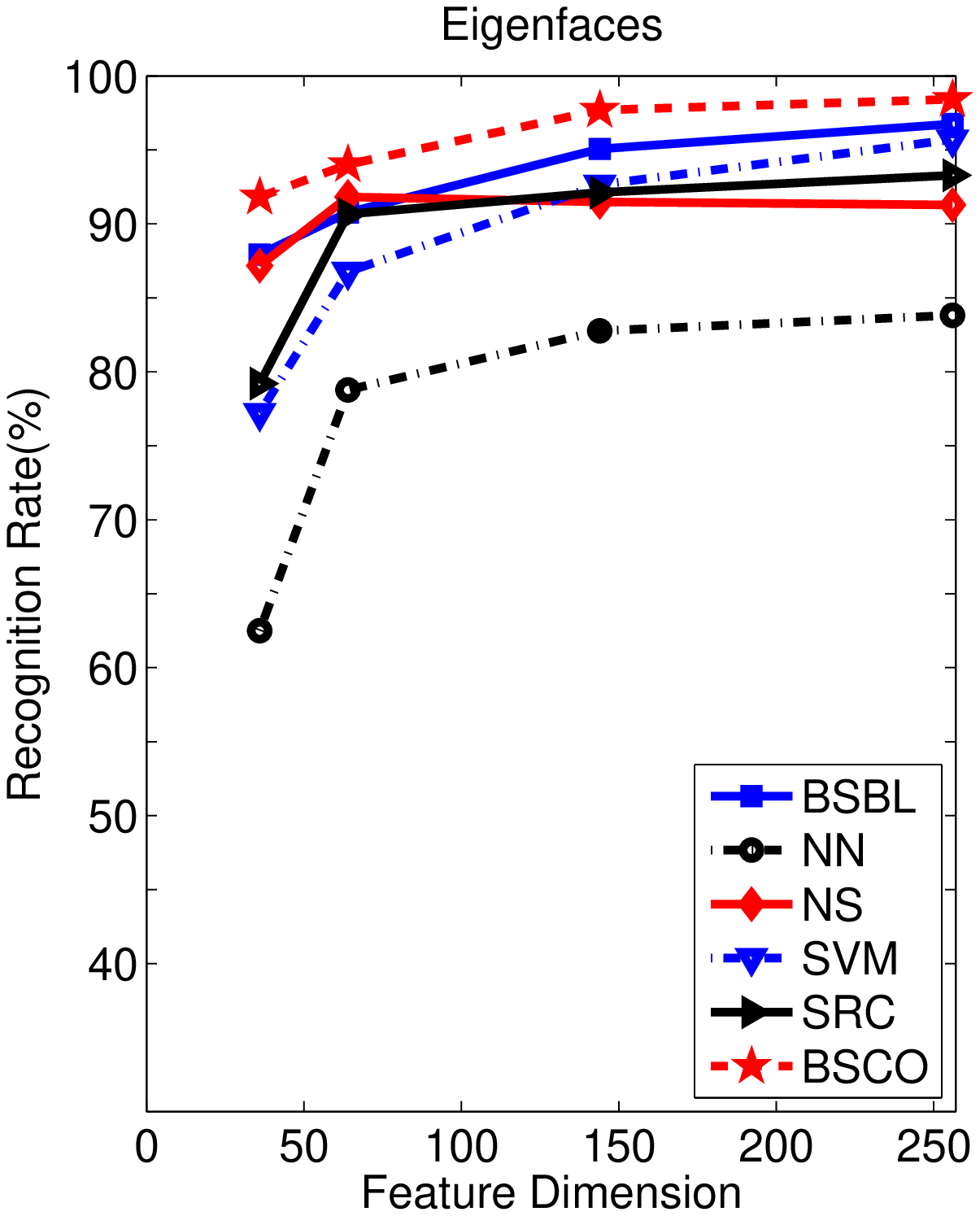,width=6cm,height=7.2cm}}
	\subfigure[]{
	\epsfig{figure=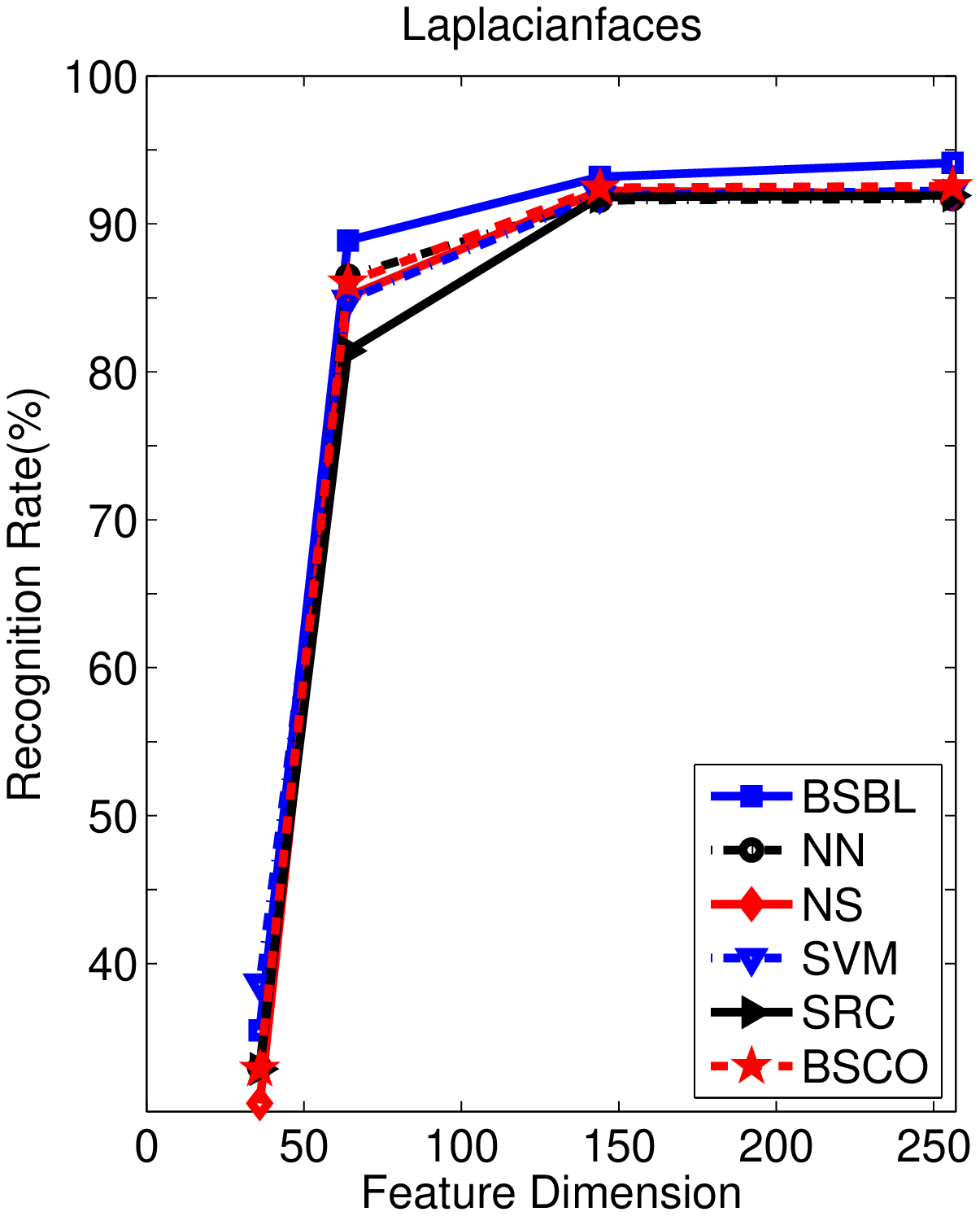,width=6cm,height=7.2cm}}
	\end{center}
	\caption{Comparison of recognition rates on CMU PIE database when using different face features. (a) Downsampling faces. (b) Eigenfaces. (c) Laplacianfaces.}
	\label{fig:exp-CMUPIE}
\end{figure}

The CMU PIE database \cite{sim2002cmu} consists of 41368 front-face images of 68 human subjects under different poses, illumination and expressions. We chose one subset(C29) which included 1632 face images of 68 subjects(24 images for each subject) for this experiment. The first subject in this subset is shown in Fig.~\ref{fig:CMUPIE-samples}, which varies in pose, illumination and expression. All the images were cropped and resized to $64 \times 64$. For each subject, we randomly selected 10 images for training, and the rest (14 images for each subject) for testing. Downsampled faces, Eigenfaces and Laplacianfaces were applied with four dimensions, i.e., 36, 64, 144 and 256. Experimental results are shown in Fig.~\ref{fig:exp-CMUPIE}.

From Fig.~\ref{fig:exp-CMUPIE}(a), we can see that sparse-representation-based classifiers usually outperformed classical ones in this dataset. Among the sparse-representation-based classifiers, BSBL and BSCO achieved higher recognition rates than SRC. For BSBL and BSCO, BSBL slightly outperformed BSCO with downsampled faces and Laplacianfaces while BSCO outperformed BSBL with Eigenfaces. Specifically, for each feature space, BSBL achieved the highest recognition rate of 95.80\% with downsampled faces and 94.12\% with Laplacianfaces while BSCO achieved 98.42\% with Eigenfaces, which was the highest one in this experiment. Nevertheless, BSBL outperformed BSCO in 8 out of 12 different combinations of dimensions and features.

\subsection{Face recognition with occlusion}
\label{subsec:withocclusion}

For the experiments on face images with occlusion, we used downsampling to reduce the size of face images and compared our method with NN \cite{duda2012pattern}, SRC \cite{wright2009robust} and BSCO \cite{elhamifar2012block}.

\subsubsection{Face recognition with pixel corruption}

\begin{figure}[!th]
	\begin{center}
	\subfigure[]{
	\epsfig{figure=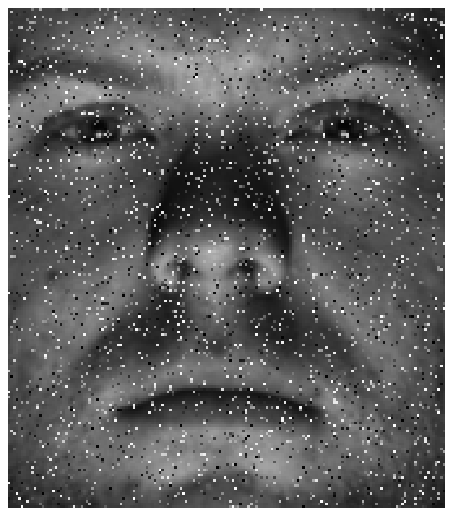,width=3.2cm,height=2.8cm}}
    \subfigure[]{
	\epsfig{figure=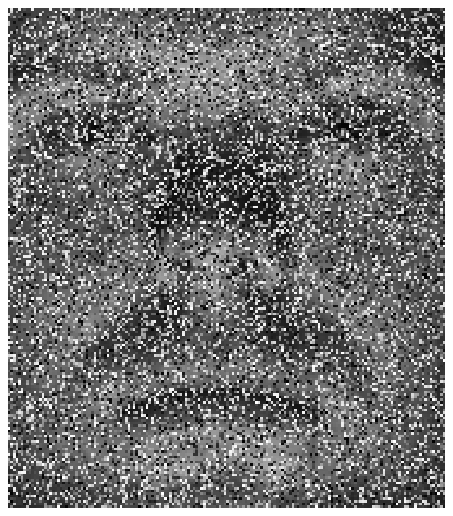,width=3.2cm,height=2.8cm}}
	\subfigure[]{
	\epsfig{figure=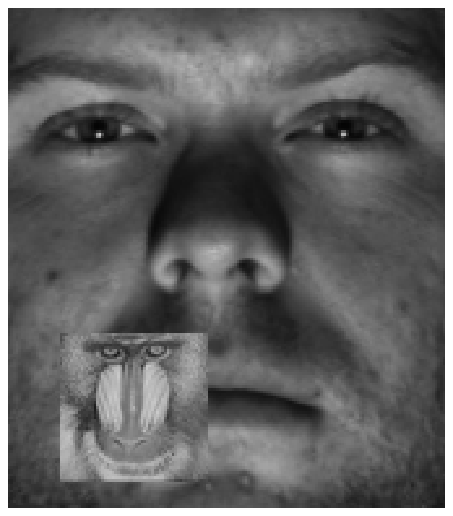,width=3.2cm,height=2.8cm}}
	\subfigure[]{
	\epsfig{figure=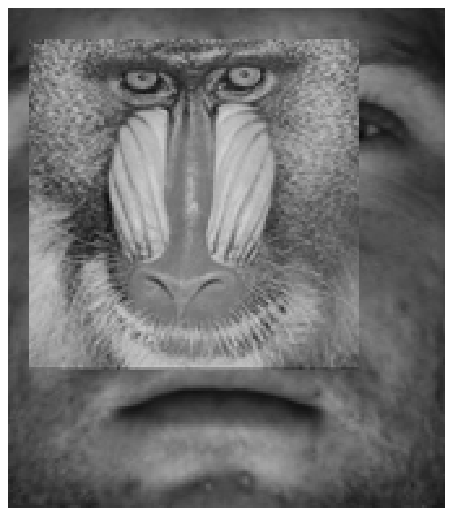,width=3.2cm,height=2.8cm}}
	\end{center}
	\caption{Corrupted and occluded face images. (a) 10\% of pixels were corrupted. (b) 50\% of pixels were corrupted. (c) 10\% of block was occluded. (d) 50\% of block was occluded.}
\label{fig:occlusion}
\end{figure}

We tested face recognition with pixel corruption on 3 subsets of the Extended Yale B database: 719 face images with normal-to-moderate lighting conditions from Subset 1 and 2 for training and 455 face images with more extreme lighting conditions from Subset 3 for testing. For each test image, we first replaced a certain percentage(0\% - 50\%) of its original pixels by uniformly distributed gray values in [0,255]. Both the gray values and the locations were random and hence unknown to the algorithms. We then downsampled all the images to the size of $6 \times 5$, $8 \times 7$, $12 \times 10$ and $24 \times 21$ respectively. Two corrupted face images were shown in Fig.~\ref{fig:occlusion}(a)-(b).

\begin{figure}[!th]
	\begin{center}
    \subfigure[]{
	\epsfig{figure=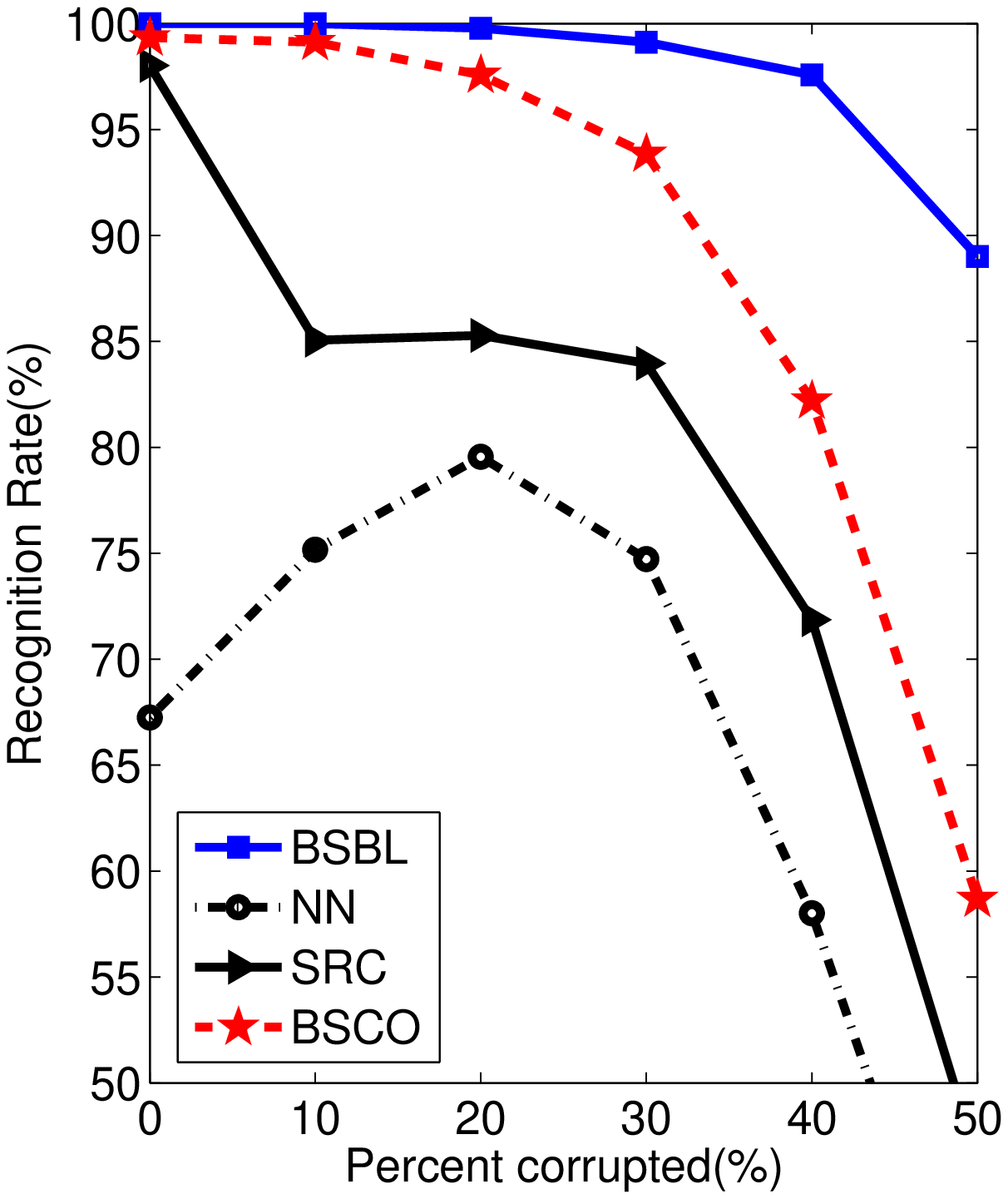,width=6cm,height=7.2cm}}
    \subfigure[]{
	\epsfig{figure=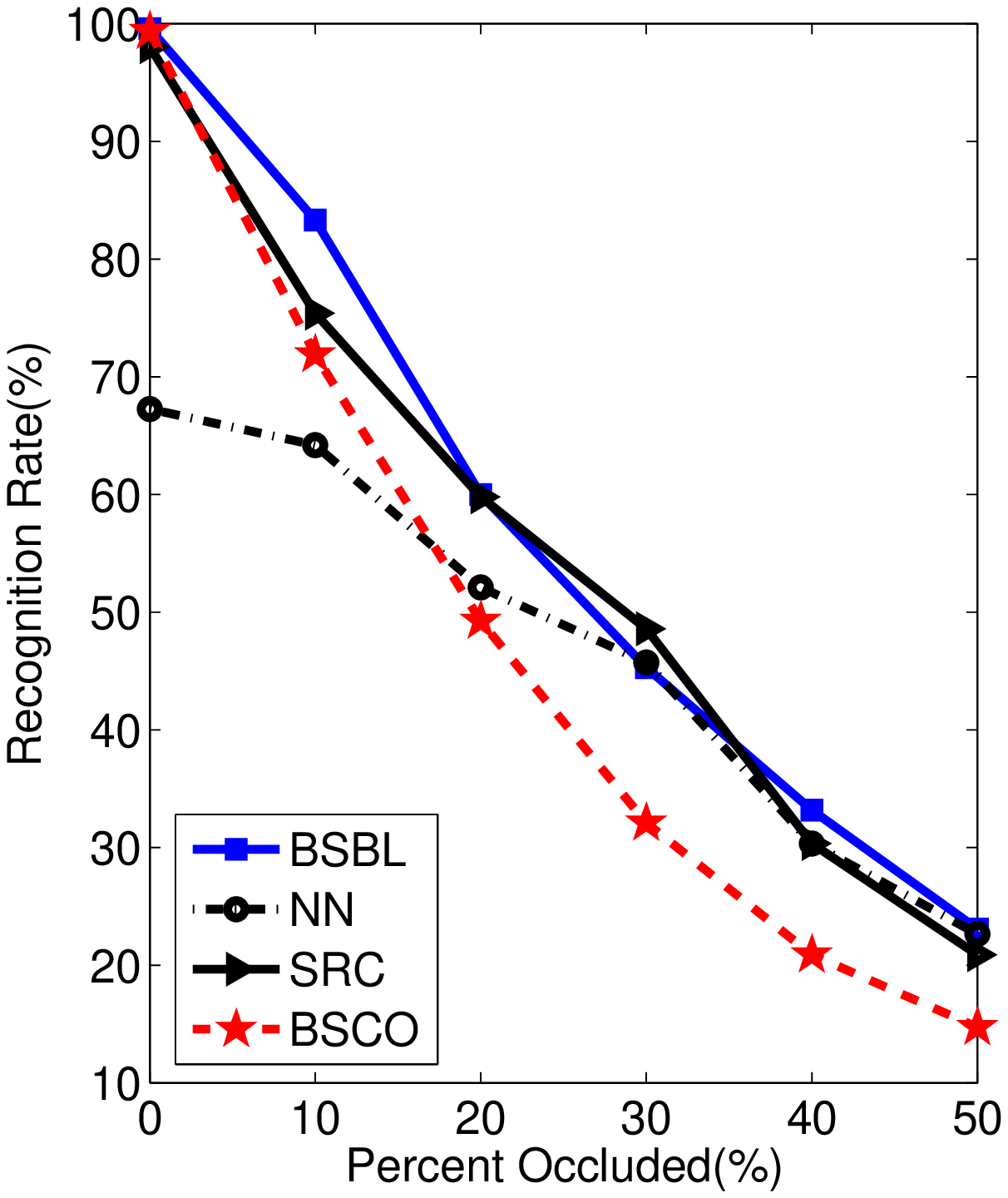,width=6cm,height=7.2cm}}
	\end{center}
	\caption{Comparison of recognition rates on faces with different percentage of (a)pixel corruption and (b)block occlusion when face dimension was $12 \times 20$.}
	\label{fig:exp-YaleB-Pixel-Corruption}
\end{figure}

Results are shown in Table~\ref{tab:pixel_corruption}. It can be seen that in almost all dimensions and corruption, BSBL achieved the highest recognition rate when compared with NN and SRC, and the performance gap between our algorithm and the compared algorithms was very large. For example, when the dimension was $12 \times 10$ and 50\% pixels were corrupted, BSBL achieved the recognition rate of 67.25\%, while SRC only had a recognition rate of 46.37\%. Meanwhile, BSBL outperformed BSCO in 17 out of 24 dimension and corruption situations. Fig.~\ref{fig:exp-YaleB-Pixel-Corruption}(a) shows the recognition rates of the four algorithms at different pixel corruption levels with the dimension of $12 \times 10$.

\begin{table}[htbp]
\caption{Recognition rate on faces with pixel corruption(\%)}
\centering
\small
\begin{tabular}{|c|c|c|c|c|c|c|c|}
\hline
\multirow{2}{*}{Method}&
\multirow{2}{*}{Dimension}&
\multicolumn{6}{|c|}{Percent corrupted(\%)} \\
\cline{3-8}
 &
 &
0&
10&
20&
30&
40&
50 \\
\hline
\multirow{4}{*}{NN}&
$6\times5$&
36.92&
42.42&
49.67&
46.15&
28.35&
14.95 \\
\cline{2-8}
 &
$8\times7$&
48.79&
54.95&
60.00&
59.34&
40.88&
20.22 \\
\cline{2-8}
 &
$12\times10$&
67.25&
75.17&
79.56&
74.73&
58.02&
35.39 \\
\cline{2-8}
 &
$24\times21$&
87.25&
93.19&
94.95&
92.53&
76.48&
56.04 \\
\hline
\multirow{4}{*}{SRC}&
$6\times5$&
54.51&
44.62&
50.55&
46.59&
32.75&
21.76 \\
\cline{2-8}
 &
$8\times7$&
82.64&
61.32&
66.59&
63.52&
49.23&
28.79 \\
\cline{2-8}
 &
$12\times10$&
98.02&
85.06&
85.28&
83.96&
71.87&
46.37 \\
\cline{2-8}
 &
$24\times21$&
100.00&
98.24&
98.24&
97.14&
92.09&
73.63 \\
\hline
\multirow{4}{*}{BSCO}&
$6\times5$&
87.48&
83.52&
63.30&
39.12&
23.30&
14.29 \\
\cline{2-8}
 &
$8\times7$&
98.02&
96.04&
90.99&
75.60&
49.45&
30.33 \\
\cline{2-8}
 &
$12\times10$&
99.34&
99.12&
97.58&
93.85&
82.20&
58.68 \\
\cline{2-8}
 &
$24\times21$&
100.00&
100.00&
99.12&
98.68&
97.14&
96.70\\
\hline
\multirow{4}{*}{BSBL}&
$6\times5$&
87.25&
85.71&
68.79&
51.43&
30.99&
19.56 \\
\cline{2-8}
 &
$8\times7$&
94.29&
92.97&
86.15&
72.53&
59.12&
39.34 \\
\cline{2-8}
 &
$12\times10$&
99.56&
99.34&
97.80&
92.31&
84.18&
67.25 \\
\cline{2-8}
 &
$24\times21$&
100.00&
100.00&
99.78&
99.12&
97.58&
89.01 \\
\hline
\end{tabular}
\label{tab:pixel_corruption}
\end{table}

\subsubsection{Face recognition with block occlusion}

In this experiment, we used the same training and testing images as those in the previous pixel corruption experiment. For each test image, we replaced a randomly located square block with an unrelated image(the baboon image in SRC \cite{wright2009robust}), which occluded 0\% - 50\% of the original testing image. We then downsampled all the images to the size of $6 \times 5$, $8 \times 7$, $12 \times 10$ and $24 \times 21$ respectively. Two occluded face images were shown in Fig.~\ref{fig:occlusion}(c)-(d).

Table~\ref{tab:block_occlusion} shows the recognition rates of NN, SRC, BSCO and BSBL on different dimensions and percentages of occlusion. Again, BSBL outperformed the compared algorithms in most cases. For example, when the occlusion percentage ranged from 10\% to 50\% and the face dimension was $12 \times 10$, BSBL achieved about 8.35\%-13.19\% higher recognition rate than BSCO, as shown in Fig.~\ref{fig:exp-YaleB-Pixel-Corruption}(b).

\begin{table}[htbp]
\caption{Recognition rate on faces with block occlusion(\%)}
\centering
\small
\begin{tabular}{|c|c|c|c|c|c|c|c|}
\hline
\multirow{2}{*}{Method}&
\multirow{2}{*}{Dimension}&
\multicolumn{6}{|c|}{Percent occluded(\%)} \\
\cline{3-8}
 &
 &
0&
10&
20&
30&
40&
50 \\
\hline
\multirow{4}{*}{NN}&
$6\times5$&
36.92&
34.29&
27.69&
24.40&
20.44&
15.17 \\
\cline{2-8}
 &
$8\times7$&
48.79&
44.84&
38.68&
32.09&
21.54&
18.46 \\
\cline{2-8}
 &
$12\times10$&
67.25&
64.18&
52.09&
45.71&
30.33&
22.64 \\
\cline{2-8}
 &
$24\times21$&
87.25&
85.50&
76.92&
67.25&
52.31&
37.14 \\
\hline
\multirow{4}{*}{SRC}&
$6\times5$&
54.51&
36.26&
28.13&
22.64&
17.36&
14.29 \\
\cline{2-8}
 &
$8\times7$&
82.64&
50.99&
39.56&
31.65&
20.66&
17.36 \\
\cline{2-8}
 &
$12\times10$&
98.02&
75.39&
59.78&
48.57&
30.33&
20.88 \\
\cline{2-8}
 &
$24\times21$&
100.00&
96.48&
89.23&
72.31&
54.29&
35.17 \\
\hline
\multirow{4}{*}{BSCO}&
$6\times5$&
87.48&
30.11&
14.51&
8.57&
5.49&
3.96 \\
\cline{2-8}
 &
$8\times7$&
98.02&
51.65&
32.53&
18.02&
13.41&
8.35 \\
\cline{2-8}
 &
$12\times10$&
99.34&
71.87&
49.23&
32.09&
20.88&
14.73 \\
\cline{2-8}
 &
$24\times21$&
100.00&
99.56&
92.97&
80.88&
63.74&
45.93 \\
\hline
\multirow{4}{*}{BSBL}&
$6\times5$&
87.25&
46.59&
28.35&
18.68&
11.65&
10.55 \\
\cline{2-8}
 &
$8\times7$&
94.29&
66.59&
40.88&
26.59&
20.22&
12.53 \\
\cline{2-8}
 &
$12\times10$&
99.56&
83.30&
60.00&
45.28&
33.19&
23.08 \\
\cline{2-8}
 &
$24\times21$&
100.00&
96.92&
92.31&
75.60&
56.48&
42.64 \\
\hline
\end{tabular}
\label{tab:block_occlusion}
\end{table}

\subsubsection{Face recognition with real face disguise}

We used a subset of AR database to test the performance of our method on face recognition with disguise. We chose 799 images of various facial expression without occlusion (i.e., the first 4 face images in each session except a corrupted image named `W-027-14.bmp') for training. We formed two separate testing sets of 200 images. The images in the first set were from the neutral expression with sunglasses (the 8th image in each session) which cover roughly 20\% of the face, while the ones in the second set were from the neutral expression with scarves (the 11th image in each session) which cover roughly 40\% of the face. All the images were resized to $9 \times 6$, $13 \times 10$, $27 \times 20$ and $42 \times 30$ respectively.

Results are shown in Table~\ref{tab:AR-disguise}. In the case of neutral expression with sunglasses, both SRC and NN acheived higher recognition rates than BSCO and BSBL. However, in the case of neutral expression with scarves, BSBL outperformed NN, SRC and BSCO significantly. Totally, BSBL achieved the highest recognition rates 72.50\% and 74.5\% with the dimensions of $27 \times 20$ and $42 \times 30$ respectively for the two testing sets, while SRC achieved the highest recognition rates 28.25\% and 44.00\% with the other two dimensions.

\begin{table}[htbp]
\caption{Recognition rate on faces with disguise(\%)}
\centering
\small
\begin{tabular}{|c|c|c|c|c|}
\hline
Method&
Dimension&
Sunglasses&
Scarves&
Total\\
\hline
\multirow{4}{*}{NN}&
$9\times6$&
35.00&
6.50&
20.75\\
\cline{2-5}
 &
$13\times10$&
48.00&
7.00&
27.50\\
\cline{2-5}
 &
$27\times20$&
65.50&
9.50&
37.50\\
\cline{2-5}
 &
$42\times30$&
68.00&
11.50&
39.75 \\
\hline
\multirow{4}{*}{SRC}&
$9\times6$&
46.50&
10.00&
\textbf{28.25}\\
\cline{2-5}
 &
$13\times10$&
72.00&
16.00&
\textbf{44.00}\\
\cline{2-5}
 &
$27\times20$&
83.00&
21.50&
52.25 \\
\cline{2-5}
 &
$42\times30$&
89.00&
37.00&
63.00 \\
\hline
\multirow{4}{*}{BSCO}&
$9\times6$&
14.50&
9.50&
12.00\\
\cline{2-5}
 &
$13\times10$&
35.00&
19.50&
27.25\\
\cline{2-5}
 &
$27\times20$&
68.00&
44.00&
56.00 \\
\cline{2-5}
 &
$42\times30$&
76.00&
50.00&
63.00 \\
\hline
\multirow{4}{*}{BSBL}&
$9\times6$&
22.00&
23.00&
22.50\\
\cline{2-5}
 &
$13\times10$&
40.50&
46.00&
43.25\\
\cline{2-5}
 &
$27\times20$&
64.00&
81.00&
\textbf{72.50} \\
\cline{2-5}
 &
$42\times30$&
65.50&
83.50&
\textbf{74.50} \\
\hline
\end{tabular}
\label{tab:AR-disguise}
\end{table}

\section{Conclusions}


Classification via sparse representation is a popular methodology in face recognition and other classification tasks. Recently it was found that using block-sparse representation, instead of the basic sparse representation, can yield better classification performance. In this paper, by introducing a recently proposed block sparse Bayesian learning (BSBL) algorithm, we showed that the BSBL is a better framework than the basic block-sparse representation framework, due to its various advantages over the latter. Experiments on common face databases confirmed that the BSBL is a promising sparse-representation-based classifier.

\section*{Acknowledgments}
This work was supported in part by the National Natural Science Foundation of China (Grant No. 60903128) and also by the Fundamental Research Funds for the Central Universities (Grant No. JBK130142 and JBK130503).





\bibliographystyle{ieeetr}
\bibliography{mpe}







\end{document}